%% file: mlsp_liu.tex
\documentclass{article}
\usepackage{amsmath,graphicx,mlspconf}
\usepackage{subfigure}
\usepackage[ruled,vlined]{algorithm2e}
\usepackage{float}
\usepackage{subfigure}
\usepackage[acronym]{glossaries}
\usepackage{threeparttable}
\usepackage[section]{placeins}

\usepackage{stfloats}

\title{Vertex-based Networks to Accelerate Path Planning Algorithms}

\name{Yuanhang Zhang \ \ \ Jundong Liu\thanks{Thanks to Ohio
    University Research Committee (OURC) for funding.}}
\address{School of Electrical Engineering and Computer Science \\ Ohio
  University }

\begin{document}
%\ninept

\maketitle

\input abs_liu


\input intro_liu
\input back_liu

\input method_liu

\input exp_liu_2

\input con_liu

% -------------------------------------------------------------------------
%\begin{figure}[htb]

%\begin{minipage}[b]{1.0\linewidth}
%  \centering
%  \centerline{\includegraphics[width=8.5cm]{image1}}
%  \vspace{2.0cm}
%  \centerline{(a) Result 1}\medskip
%\end{minipage}
%
%\begin{minipage}[b]{.48\linewidth}
%  \centering
%  \centerline{\includegraphics[width=4.0cm]{image3}}
%  \vspace{1.5cm}
%  \centerline{(b) Results 3}\medskip
%\end{minipage}
%\hfill
%\begin{minipage}[b]{0.48\linewidth}
%  \centering
%  \centerline{\includegraphics[width=4.0cm]{image4}}
%  \vspace{1.5cm}
%  \centerline{(c) Result 4}\medskip
%\end{minipage}
%
%\caption{Example of placing a figure with experimental results.}
%\label{fig:res}
%
%\end{figure}

\small
\bibliographystyle{IEEEbib}
\bibliography{strings,refs,one_for_all}

\end{document}

%% file: abs_liu.tex
\begin{abstract}

  Path planning plays a crucial role in various autonomy applications,
  and RRT* is one of the leading solutions in this field. In this
  paper, we propose the utilization of vertex-based networks to
  enhance the sampling process of RRT*, leading to more efficient path
  planning.

  Our approach focuses on critical vertices along the optimal paths,
  which provide essential yet sparser abstractions of the paths. We
  employ focal loss to address the associated data imbalance issue,
  and explore different masking configurations to determine practical
  tradeoffs in system performance. Through experiments conducted on
  randomly generated floor maps, our solutions demonstrate significant
  speed improvements, achieving over a 400\% enhancement compared to the
  baseline model.

%In this article, we present a novel approach to improve the learned
%heuristic for path planning algorithms. Rather than using the entire
%optimal path line segment as the objective, we focus solely on the
%turning points of the optimal path. This modification significantly
%enhances the speed of the path planning algorithms. To address the
%imbalanced data problem caused by the vertex objective, we employ the
%Focal Loss technique in our deep learning model. Additionally, we
%introduce a mask to the neural network sampler, which boosts the
%sampling probability of the vertices in the optimal path. This further
%accelerates the convergence speed to the optimal path.

%Overall, our approach provides a more efficient method for learning
%heuristics, which is important for future machine learning
%applications in path planning problems.

\end{abstract}

%% file: intro_liu.tex
\section{Introduction}

Path planning aims to determine a feasible route for an autonomous
agent to travel from a starting point to a target location within an
environment while avoiding obstacles. This process has a wide range
of applications across various domains.
%In robotics, for instance, path planning facilitates the navigation
%of mobile robots in unknown environments. Self-driving cars rely on
%path planning algorithms to identify the safest and most efficient
%routes to their destinations. Path planning is also crucial in
%aerospace for aircraft to determine the most secure and
%fuel-efficient flight paths.  Overall,
The common goal of path planning is to discover a route that
is safe, efficient, and smooth.

%Path planning is the process of finding a feasible path for an
%autonomous agent to travel from a starting point to a goal in an
%environment while avoiding obstacles.  It has a wide range of
%application in various areas. For example, in robotics, path planning
%is used for navigation of mobile robots in unknown
%environments. Self-driving cars rely on path planning algorithms to
%find the most efficient and safe route to their destinations. Path
%planning is also crucial in aerospace for aircrafts to determine the
%safest and most fuel-efficient flight path.  The common goal of path
%planning is to find a path that is safe, efficient, and smooth.

% \subsection{Area Overview: path planning algorithms}

Traditional path planning algorithms can be grouped into two primary
categories: grid search-based and sampling-based. Among grid-search
algorithms, the A* algorithm \cite{hart1968formal} is one of the most
prominent solutions, capable of guaranteeing the finding of an optimal
path if one exists; however, it may encounter difficulties in
high-dimensional state spaces. Sampling-based algorithms, such as {\it
  Rapid Random-exploring Trees} (RRT) \cite{rrt} and {\it Optimal
  Rapid Random-exploring Trees} (RRT*) \cite{rrtstar}, operate through
randomly selecting states from the state space, rather than
investigating all possible states, therefore speeding up the
exploration process.  RRT uniformly sample states within the state
space while gradually building a tree structure of these states. RRT*
enhances RRT by reorganizing the tree, granting it probabilistic
completeness and asymptotic optimality.

Recently, machine learning-based approaches have been proposed to
address the intricate challenges associated with path planning.  These
approaches can generally be categorized into supervised learning (SL)
and reinforcement learning (RL) methods. SL-based solutions perform
perception and decision-making simultaneously, predicting control
policies directly from raw input images
\cite{qureshi2019motion}. RL-based methods, on the other hand, rely on
human-designed reward functions, allowing learning agents to explore
policies through trial and error \cite{tamar2016value}. While
promising, learning-based path planning solutions often lack
theoretical guarantees on performance. Moreover, SL requires annotated
data, which can be difficult or expensive to acquire.

The latest RRT-based solutions, including informed RRT*
\cite{gammell2014informed} and connect RRT \cite{gammell2015batch},
while using traditional strategies, are regarded as the
state-of-the-art solutions in the field. This can be attributed to
their flexibility in handling changes in the environment and their
capability of to navigate high-dimensional state spaces. Moreover,
RRT* has the guarantee of asymptotic optimality and probabilistic
completeness, which ensures that the solution achieves optimality
under specific conditions.

However, RRT* solutions suffer sensitivity to the initial solution and
slow convergence to the optimal solution. To overcome these
limitations, several network-based solutions have been proposed to
speech up the sampling process. Trained on optimal paths, Neural RRT*
\cite{wang2020neural} and Motion Planning Networks
\cite{chen2019motion} predict the probability distribution of the path
to achieving faster samplings. Neural Informed RRT*
\cite{xie2019neural} guides RRT* tree expansion using an
offline-trained neural network during online planning.

Although considerable speed-ups have been demonstrated in comparison
to the original RRT* algorithm, the aforementioned acceleration
networks
%commonly rely on paths as the basis and target for reducing the search
%space.
commonly take the entire A* search space as the target area and
estimate probabilities based on the proximity to optimal paths.
% predict, which limits the rate of search speedup.
Moreover, they may struggle with highly dynamic environments or those
with rapidly changing obstacles, as the planning may become quickly
outdated.

%In this paper, we aim to provide a remedy to further RRT* speedup
%through the reduction of the target search space. Our approach is to
%switch the search from path neighborhood to vertices (turning
%points), which would provide a higher level of abstraction of the
%optimal paths, leading to enhanced speed-up and robustness.
% switching from search areas to vertices (turning points)
%of the paths.
%
% Paths, however, can be further approximated using more
% elementary components, such as line segments. In this paper, we
% further the speed-up by adopting vertices as the target entities,
%significantly reducing the search space for the network.
%

\vspace{0.5in}

In this paper, we propose to enhance the speed-up of the sampling
process by shifting the target areas from the neighborhood of optimal
paths to that of vertices (corners or turning points). Our design is
based on the rationale that critical vertex points in the optimal
paths provide an insightful and adequate abstraction of the paths,
while requiring much less space. Focusing on vertices, however,
results in a side effect that the training data would be highly
imbalanced. We address this issue using focal loss \cite{lin2017focal}
in this work. We also explore different thresholding setups for the
network outputs to examine the system tradeoffs in performance.

%% file: back_liu.tex
\section{Background}

%Rapidly-Exploring Random Trees (RRT) is one family of sampling based
%path planning algorithms. RRT \cite{rrt} was first introduced by
%LaValle in 1998. It incrementally build a tree structure. At the first
%step, only the start state is included in the vertices set, the edge
%set is empty. In the next part, the RRT sample state $x_{\rm rand}$
% from the search space uniformly, this will gurantee the entire search
% space can be covered. Then the nearest vertex $x_{\rm nearest}$ in the
% graph $G$ will be selected. The Steer function control the growth
%length of the tree, a $x_{\rm new}$ lies between $x_{\rm nearest}$ and
%$x_{\rm x_rand}$ will be returned. This $x_{\rm new}$ will be added to
%the vertices set $V$ and the edge $(x_{\rm nearest},x_{\rm x_rand})$
%is added to the edge set $E$. This iteration will process until
%certain number of iterations are met. The RRT* will return the graph
%$G$.

Rapidly-Exploring Random Trees (RRTs) comprise a family of path
planning algorithms that depend on incremental sampling. The RRT
algorithm \cite{rrt} starts with a single-vertex tree that represents
the initial state and has no edges. Over each iteration, the algorithm
generates a state $x_\textrm{rand}$ from a uniform sampling of the
search space and tries to link it to the nearest vertex
$x_\textrm{nearest}$ in the tree. If this linkage is feasible, the
Steering function manipulates $x_\textrm{rand}$ to produce
$x_\textrm{new}$. The new state $x_\textrm{new}$ and new edge
$(x_\textrm{nearest}, x_\textrm{new})$ are then added to the growing
tree.

%is then added to the set of vertices, and the edge
%(xnearest, xnew) is added to the set of edges.
%The algorithm concludes either when the tree incorporates a
%state in the goal region or the number of iterations exceeds a
%specified threshold.

%The optimal Rapidly-Exploring Random Trees (RRT*) \cite{rrtstar} is
%the optimal version of RRT. It has two additional procedures,
%${\rm Extend}(G,x_{\rm new})$ and ${\rm Rewire}(G)$. ${\rm
%  Extend}(G,x_{\rm new})$ will search near vertices within a radius,
%choose the one with minimum cost as the parent. The ${\rm Rewire}(G)$
%will reconnect the vertices in that radius if the path through $x_{\rm
%  new}$ provides a shorter path. Detail of the RRT* algorithms is
%illustrated in the Algorithm \ref{algorithm.rrtstar}. The ${\rm
%  ObstacleFree}(x_{\rm nearest},x_{\rm new})$ function determines
%whether the line segment between $x_{\rm nearest}$ and $x_{\rm new}$
%exist any obstacles. The RRT* is asymptotic optimal, as the sampling
% iterations go to infinity, the path converges to the optimal path.

The RRT* algorithm \cite{rrtstar} introduces two additional procedures:
${\rm Extend}(G,x_{\rm new})$ function and the ${\rm Rewire}(G)$
process.  During the ${\rm Extend}$ procedure, RRT* searches for
optimal parent vertices around $x_\textrm{new}$ within a certain
radius. After integrating $x_\textrm{new}$ into the tree, RRT* rewires
neighbor vertices to assess whether a path through $x_\textrm{new}$
can provide a lower cost than the current path. The procedure of the
RRT* algorithm is illustrated in Algorithm \ref{algorithm.rrtstar}.
$\textrm{ObstacleFree}(x_\textrm{nearest},x_\textrm{new})$ function
determines if the line segment connecting $x_\textrm{nearest}$ and
$x_\textrm{new}$ is obstacle-free.  RRT* is asymptotically optimal,
which means that as the sampling iterations approach infinity, the
path converges to the optimal path.

\begin{algorithm}
	\caption{RRT*}%算法名字
	\label{algorithm.rrtstar}
	\LinesNumbered %要求显示行号
	\KwIn{$x_{\rm init},x_{\rm goal},Map$}%输入参数
	\KwOut{$G=(V,E)$}%输出
	%some description\; %\;用于换行
	$V \leftarrow \{x_{\rm init}\}; E \leftarrow \emptyset$ \;
	\For{$i=1,\cdots,n$}{
		$x_{\rm rand} \leftarrow$ UniformSample() \;
		$x_{\rm nearest} \leftarrow$ Nearest($G=(V,E),x_{\rm rand})$ \;
		$x_{\rm new} \leftarrow$ Steer($x_{\rm nearest},x_{\rm rand})$ \;
		\If{ ${\rm ObstacleFree}(x_{\rm nearest},x_{\rm new})$ }{
			Extend($G,x_{\rm new}$) \;
			Rewire($G$) \;
		}
	}
	\Return{$G=(V,E)$} \;
	
\end{algorithm}

%Informed RRT*
%\cite{gammell2014informed} proposed by Gammell \textit{et al.} in 2014
%reduces the sampling region by an ellipsoidal subset. This
%modification is based on RRT*, therefore it retains the nice
%properties of RRT*.  Some advanced heuristic has been carried out to
%guild the search more intelligently \cite {aitstar} \cite{bitstar}
%\cite{abitstar}.

%Neural RRT* \cite{wang2020neural} learns a neural network to sample
%state from the the search space. Instead of the uniform sampling, its
%sampling step has 50\% probability to sampling state from the learned
%neural network, which speed up the search efficiency of RRT*. Our
%proposed Vertex Net RRT* inherited its algorithm structure,
%illustrated in Algorithm \ref{algorithm.vertexnetrrtstar}.

The Neural RRT* algorithm \cite{wang2020neural} trains a CNN model on
successful path planning cases to generate a nonuniform sampling
distribution. %The model is trained using A* algorithm to generate a
%dataset consisting of map information and optimal path.
For a given task, the trained network can predict the probability
distribution of the optimal path for on the map, which can guide and
speed up the sampling process.

%% file: method_liu.tex
\section{Method}

%Neural RRT* and its variants use the predicted probability
%distribution of the path to accelerate the sampling process. However,
%these acceleration networks usually consider the entire A* search
%space as the target area and estimate probabilities based on optimal
%paths.
%
%
%As mentioned in the {\it Introduction} section, we propose to further
%the speed-up of the Neural RRT* by shifting the target areas of the
%neurla networks from the neighborhoods of optimal paths to those of
%key vertices (turning points).
%We model our vertex extraction step as a semantic segmentation
%problem.

In this work, we propose a method to improve the Neural RRT* category
by redirecting the sampling guidance from the neighborhoods of optimal
paths to key vertices. To achieve this, we train a neural network
called {\it VertexNet}, and subsequently integrate it with the RRT*
algorithm.

%
%To enable the heuristic function to learn from data, we trained a
%neural network for this purpose. We also improved the objective
%function used for model training by focusing only on the turning
%points of the optimal path, rather than the entire path segment. The
%rationale behind this approach is that only the vertex points of the
%optimal path are necessary for constructing it, while the line
%segments between these points are functionless.
%However, this advanced
%objective function results in highly imbalanced data, which we address
%by using focal loss \cite{lin2017focal}. To increase the probability
%of sampling vertices of the optimal path, we designed a mask for the
%sampling probability distribution based on the predicted pixel
%probability $p\in[0,1]$ of the neural network. This efficient approach
%accelerates the convergence speed towards the optimal solution. Since
%we inherited the RRT* algorithm structure, the Vertex Net RRT*
%maintains the probabilistic completeness and asymptotic optimality
%properties.
%\vspace{-6pt}

%\subsection{Deep Convolutional Neural Networks Model}

%\subsection{VertexNet to Predict Vertex-ness Probablities}
\subsection{VertexNet}
\begin{figure}[H]
	\centering
        \includegraphics[width=\linewidth,angle=0]{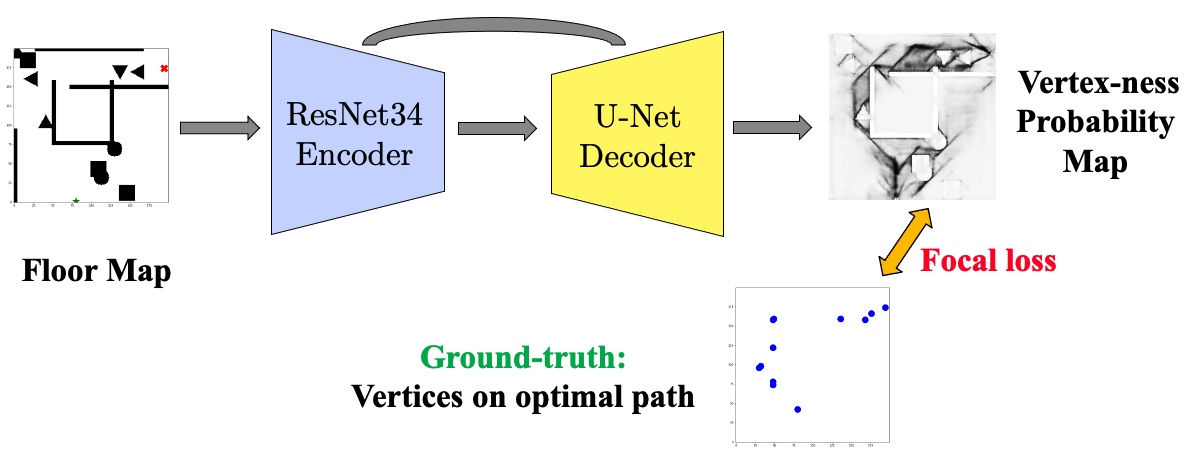}
	\caption{Illustration of the proposed VertexNet. Best view on
          screen. }
	\label{fig.model}
\end{figure}

Our VertexNet is designed to predict the likelihood of each pixel
being a vertex on the optimal path, which we refer to as {\it
  vertex-ness}.
%In this paper, we interchangeably use the terms vertex, corner, and
%turning-point.
We approach this task as an image mapping problem and
address it using a fully convolutional network, as depicted in
Fig.~\ref{fig.model}.  The input to VertexNet consists of an RGB image
representing a floor map, where obstacles, source, and target points
are differentiated by distinct colors.
The ground-truth is a corresponding vertex map extracted from the A*
optimal path. The output of VertexNet is a vertex-ness map, which will
subsequently be integrated into the RRT* algorithm to guide the
sampling process.

%Our proposed image-mapping network is
VertexNet is modified from the U-Net \cite{unet},
% where the modifications are centered on
primarily by adopting ResNet34, a residual network \cite{he2016deep}
as the backbone for the encoder. This modification aims to enhance the
network's ability to capture important features from the input images
and facilitate effective training. The updated encoder consists of
basic blocks as described in \cite{he2016deep}, each of which contains
two 2-dimensional convolutional layers, two batch normalization
layers, and one Rectified Linear Unit (ReLU) activation.
%\cite{relu}.
In total, our network has 54 weighted layers and 41,221,168 parameters,
among which, 19,953,520 are trainable.  The network takes floor map
images of size $200 \times 200$ as inputs.
%In our dataset, the input is
%of size 200x200 pixels, although this U-Net model structure can work
%with any input shape. The total number of

%The first convolutional layer is followed by a batch
%normalization layer, then a ReLU activation is applied after it. In
%total, our network has 54 weighted layers.

%In our dataset, the input is of size 200x200 pixels, although this
%U-Net model structure can work with any input shape.

%We propose a convolutional based deep neural networks model, which is
%illustrated in figure \ref{fig.model}. The U-Net \cite{unet} is the
%overall model structure. The residual networks \cite{he2016deep}
%ResNet34 works as the backbone for the encoder part of the U-Net.  The
%model consists of 54 weighted layers. In our dataset, the input is of
%size 200x200 pixels, although this U-Net model structure can work with
%any input shape. The encoder consists of basic blocks described by
%\cite{he2016deep}. Each basic block consists of two 2 dimensional
%convolutional layers, two batch normalization layers, and one
%Rectified Linear Unit (ReLU) activation \cite{relu}. The first
%convolutional layer is followed by a batch normalization layer, then a
%ReLU activation is applied after it.

\subsubsection{Ground Truth and Training Objective}

%\begin{itemize}
%\item The ground-truth of our network are binary images generated from
%  the optimal paths.
%\item We set the pixel intensities of a  small neighborhood of chosen vertices to 1;
%  and other are 0s.
%\item The vertices are determined based on a corner-ness
%  consideration.
%\item There are many different ways to define corners. In this work,
%  as the opotimal paths go through pixel domain (i.e., pixel grids),
%  our focus is eliminate non-corner cases.  More specifically, we
%  eliminate points on straight lines (as shown on fig 3.1), points on
%  45-degree zigzag lines, and points on 22.5-degree zigzag lines,
%  shown on fig. 3.3. The latter two cases can be counted staight lines
%  in a larger scale.
%  \item This design can translate into a predict as follows. 
%\end{itemize}

The ground-truth images in our work are generated following a
three-step process. In the first step, we employ the A* algorithm to
determine the optimal path for a floor map. Next, we extract a number
of vertex points on the optimal path based their vertex-ness (being
corners or turning points). Finally, we create the ground-truth images
by setting the pixels
%in a small neighborhood
of the selected vertices to 0s, while the remaining pixels are set to
1s.

%The ground truth for our network consists of binary vertex images
%generate from the optimal paths. Pixels in a small neighborhood
%surrounding selected vertices have intensity of 1, while the while the
%remaining pixels are set to 0. The vertices are all on the optimal
%path, and they are determined based on their ``corner-ness'' (being
%corners along the path).
%which involves
%different ways of defining corners.
%

\begin{figure}[htb]
        \centering \subfigure[0-degree and 90-degree lines.]{
                \label{fig.map.map1}
                \includegraphics[width=0.45\linewidth]{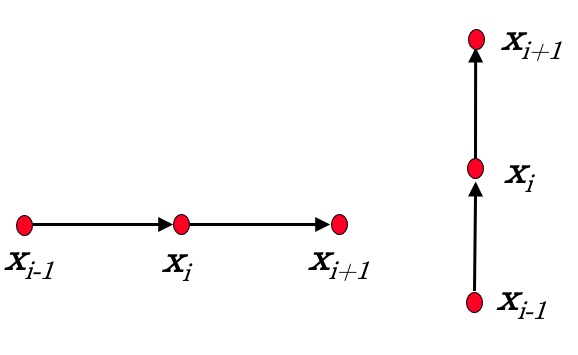}
        }
        \subfigure[45-degree straight lines.]{
          \label{fig.map.map1}
          \includegraphics[width=0.45\linewidth]{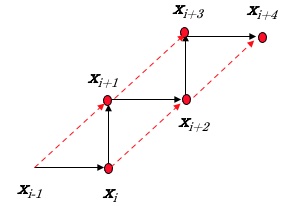}
        }
        \subfigure[22.5-degree straight  lines. ]{
          \label{fig.map.map1}
          \includegraphics[width=0.5\linewidth]{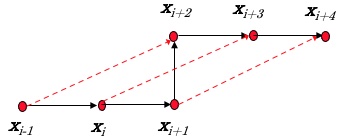}
        }
        \caption{Three cases of straight lines within the pixel
          domain. }
        \label{fig:ver_123}
        \end{figure}

Vertices are chosen among the pixels on the path based on specific
criteria. Specifically, only the end-points of line segments are
considered as vertices, while intermediate points are excluded. Since
the paths are generated within discrete image domains, the
identification of straight lines becomes crucial.  Straight lines can
consist of line segments with consistent directions, as in
Fig.~\ref{fig:ver_123}(a).  The red dash-lines in
Fig.~\ref{fig:ver_123}(b) could also be straight under a continuous
domain, as their directions are not changed.
Fig.~\ref{fig:ver_123}(c) shows a similar case where we look ahead for
three steps. It should be noted that the points $x_i$ in all three
cases should not be classified as turning points.

%Specifically, we exclude points that lie on straight lines, as shown
% in Fig.~\ref{fig:ver_123}(a), as well
%as points along 45-degree zigzag lines (red dash-lines in
%Fig.~\ref{fig:ver_123}(b)) and 22.5-degree zigzag lines (red
%dish-lines in Fig.~\ref{fig:ver_123}(c)).  It should be noted that, in
%continuous domains, the latter two cases are actually straight lines.

%In our study, the optimal paths are defined in the discrete image
%domains, through either 4-connection of 8-connection.  Our main focus
%is to exclude non-corner cases. Specifically, we eliminate points that
%lie on straight lines, as shown in Fig.~\ref{fig:ver_123}(a), as well
%as points along 45-degree zigzag lines (red dash-lines in
%Fig.~\ref{fig:ver_123}(b)) and 22.5-degree zigzag lines (red
%dish-lines in Fig.~\ref{fig:ver_123}(c)).  It should be noted that, in
%continuous domains, the latter two cases are actually straight lines.

Let $X_\textrm{path} = \{x_0,x_1,\cdots,x_T\}$ be a path going through
a sequence of positions within an image domain. To determine whether a
point $x \in X_\textrm{path}$ should be considered as a turning point,
we examine if there is a change in direction for any of the three step
sizes in Fig.~\ref{fig:ver_123}. In essence, we define a vertex as
follows:

\begin{equation*}
%\resizebox{.9\hsize}{!}{
%	$X_{vertex}=\{x\in X_{path} \mid \Delta(x_i-x_{i-1}) \land \Delta{(x_i-x_{i-2}) %\land \Delta{(x_i-x_{i-3})}}\}$
%}
\begin{aligned}
	X_\textrm{vertex}=\{x\in X_\textrm{path} \mid &\Delta(x_i-x_{i-1}) \land \Delta(x_i-x_{i-2})\\
	&\land \Delta(x_i-x_{i-3})\}
\end{aligned}
\end{equation*}
where $\Delta$ denotes the presence of a change in direction with a
certain step size.

%We define the $x \in X_{path}$ as the vertex point when the change of
%direction occurred both by a difference of period 1, period 2, and
%period 3, which restrict the condition for our map with 8 directions.

%To generate the ground truth for model training, we follow a
%three-stage process.  In the first stage, we employ the A* algorithm
%to determine the optimal path. Next, in the second stage, we extract
%the vertex points from the optimal path using our vertex point
%definition. Finally, in the third stage, we create the ground truth
%using the extracted vertex points. We treat the map as a binary mask,
%where the vertex points and the goal location are assigned one class,
%while all other pixels are assigned the other class. Consequently, the
%prediction target of our model will be the vertex points along the
%optimal path and the goal point. This approach drastically reduces the
%sampling space.

%To generate the ground truth for model training. In the first stage,
%we apply the A* algorithm to get the optimal path. For the second
%stage, we use our vertex point definition to extract the vertex points
%from the optimal path. Then, in the third stage, we can create the
%ground truth from the vertex points. The map will be treated as the
%binary mask, the vertex points, and the goal of the map will be one
%class, all the other pixels will be the other class. Therefore, the
%prediction target of our model will be the vertex points of the
%optimal path and the goal point, which will reduce the sampling space
%drastically.

\subsubsection{Loss Function}

As very few pixels in each ground-truth image are selected as
vertices, the ground-truth images tend to be predominately blank as
vast majority (over 99\%)  of the pixels have intensity of 1s.
%
%In our work,
%the vertices only take several pixels in the image, most (over 99\%)
%pixels are foreground.
%
This intensity imbalance would create a challenge for image mapping
problems. To tackle this issue, we adopt the focal loss
\cite{lin2017focal} as the objective function for our VertexNet.
\begin{equation*}
	{\rm FL}(p_{\rm t}) = -(1-p_{\rm t})^\gamma \log(p_{\rm t})
\end{equation*}

Compared to the cross-entropy loss, the focal loss adds a modulating
factor $ (1-p_{\rm t})^\gamma $.  This factor assigns varying weights
to samples based on their difficulty of classification. Easy samples,
which are more likely to be correctly classified, receive reduced
weights, while harder samples are assigned higher weights. In our
dataset, as the non-vertex pixels constitute the majority, they are
categorized as easy samples, resulting in reduced contributions
through the modulating factor. The focusing hyperparameter $\gamma$ is
adjustable, and based on empirical observations, we set it to $2$ in
our experiments.

%
%This modulating factor assigns
%different weights to easy-classified samples and hard-classified
%samples. For easy samples, the weights should be reduced and vice
%versa for hard samples. In our data, as non-vertex pixels acount for
%vast majority, they are regarded as easy samples and given reduced
%contrbutions through the modulating factor.
%
%If the probability $p_{\rm t}$ is numerically large, it is
%an easily classified example, the loss it contributed to the training
%will less weighted. For a hard classified example, the probability $
%(1-p_{\rm t})^\gamma $ will be low, it contributes more loss relative
%to the hard examples.
%The focusing hyperparameter $\gamma$ is tunable and, based on
%empirical observations, we set it to 2 in our experiments.
%The $\gamma$ is a tunable focusing hyperparameter and set to 2 in our
%experiments.  used in our setting.

\subsection{VertexNet RRT*}	

%The Vertex Net RRT* adopts the RRT* algorithm structure to maintain
%the probabilistic complete and asymptotical optimal properties. The
%process of Vertex Net RRT* is illustrated in
%Algorithm~\ref{algorithm.vertexnetrrtstar}.

After VertexNet is trained, it is integrated into the RRT* algorithm
to enhance the sampling process. The integration process follows a
similar design to that of Neural RRT* \cite{wang2020neural}.
Taking a floor map as the input, the non-uniform sampler generated by
VertexNet works together with the uniform sampler from RRT* to make
informed sampling decisions. VertexNet plays a crucial role by
providing probabilistic guidance for selecting the next sampling
point, thus improving the overall sampling efficiency. Meanwhile, the
uniform sampler ensures that the integrated algorithm retains its
original properties of probabilistic completeness and asymptotic
optimality.

The resulting algorithm is referred to as VertexNet RRT*, as outlined
in Algorithm~\ref{algorithm.vertexnetrrtstar}. Each sampler is
assigned a $50\%$ probability of being invoked, which is determined by
the $\textrm{Rand()}$ function. The Nearest function, Extend, Rewire
and Steer function are all the same as in the original RRT* algorithm.

\begin{algorithm}
	%	\setstretch{1}
%	\singlespace
	\caption{VertexNet RRT*}%算法名字
	\label{algorithm.vertexnetrrtstar}
	\LinesNumbered %要求显示行号
	\KwIn{$x_{\rm init},x_{\rm goal},Map,{\rm VetexNet}$}%输入参数
	\KwOut{$G=(V,E)$}%输出
	%some description\; %\;用于换行
	$\mathcal{O} = {\rm VertexNet}(Map,x_{\rm init},x_{\rm goal})$ \;
	$V \leftarrow \{x_{\rm init}\}; E \leftarrow \emptyset$ \;
	\For{$i=1,\cdots,n$}{
		\eIf{${\rm Rand}() > 0.5$}{
			$x_{\rm rand} \leftarrow$ VertexNetSample($\mathcal{O}$) \;
			
		}{
			$x_{\rm rand} \leftarrow$ UniformSample() \;
		}
		$x_{\rm nearest} \leftarrow$ Nearest($G=(V,E),x_{\rm rand})$ \;
		$x_{\rm new} \leftarrow$ Steer($x_{\rm nearest},x_{\rm rand})$ \;
		\If{ ${\rm ObstacleFree}(x_{\rm nearest},x_{\rm new})$ }{
			Extend($G,x_{\rm new}$) \;
			Rewire($G$) \;
		}
	}
	\Return{$G=(V,E)$} \;
\end{algorithm}

%The input for our Vertex
%Net RRT* is a map associated with an initial state $x_{init}$ and a
%goal state $x_{goal}$. Another component is the VertexNet model, which
%has been trained on 1,008,000 maps to gain experience. The output
%inherited the RRT* output, which is a tree. The input map
%will be fed into the VertexNet to get the probability distribution for
%each pixel to be a vertex point of the optimal path.

%For each iteration of the algorithm, a uniform sampling in $[0,1]$
%will be executed, when the realization is above 0.5, the random sample
%for this iteration will be generated from the VertexNet sampler,
%otherwise, it uniform samples the random state. These will assign
%equal probability to both samplers, which keep the probabilistic
%completeness and the asymptotical optimality of RRT*, hence,
%our
%proposed Vertex Net RRT* is also an optimal planning algorithm.
%The Nearest function is inside the Extend procedure, which returns the
%node in the tree that is closest to $x_{rand}$.  The Steer function
%will control the step distance for tree growing, which is
%hyperparameter $\eta$. The $x_{new}$ lies between the $x_{nearest}$
%and $x_{rand}$.  The ObstacleFree function checks whether any obstacle
%lies between $x_{nearest}$ and $x_{new}$.

\subsubsection{Masked VertexNet RRT*}
%We have developed a masked version of the Vertex Net RRT* algorithm,
%in which a mask is applied to the neural network output representing
%the sampling probability distribution. This mask sets the predicted
%probabilities below a certain threshold value, denoted by $\tau$, to
%zero. Without the mask, pixels with probabilities below 50\% are
%unlikely to be vertex points. However, in the sampling process, some
%probability is still assigned to these pixels. With the mask applied,
%probabilities below the threshold value are not considered, resulting
%in a higher likelihood of sampling vertex points.
%\vspace{-0em}

To explore different configurations, we develop a masked version of
the VertexNet RRT* algorithm. In this variant, a mask is applied to
the sampling probability distribution generated by VertexNet. The mask
functions by setting probabilities below a specific threshold value,
denoted as $\tau$, to zero. In the original VertexNet RRT* algorithm,
even pixels with low probabilities have a chance of being chosen as
vertex points. However, with the applied mask, probabilities below the
threshold value are disregarded, leading a higher likelihood of
sampling actual vertex points.

%% file: exp_liu_2.tex
\section{Experiments}

%\subsection{Data Generation and Network traing}

\begin{figure}[htb]
	\centering
	\subfigure[Map 1]{
		\label{fig.map.map1}
		\includegraphics[width=0.3\linewidth]{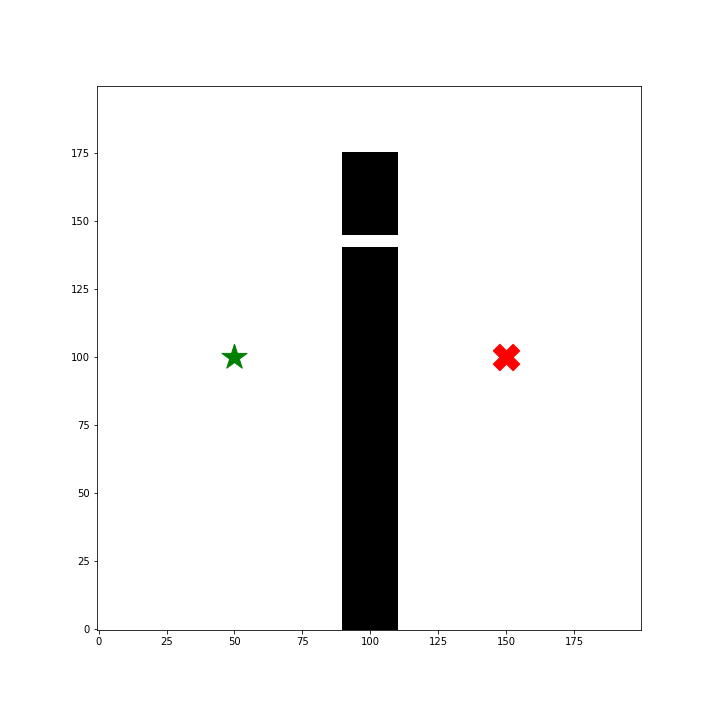}
	}
	\subfigure[Map 2]{
		\label{fig.map.map2}
		\includegraphics[width=0.3\linewidth]{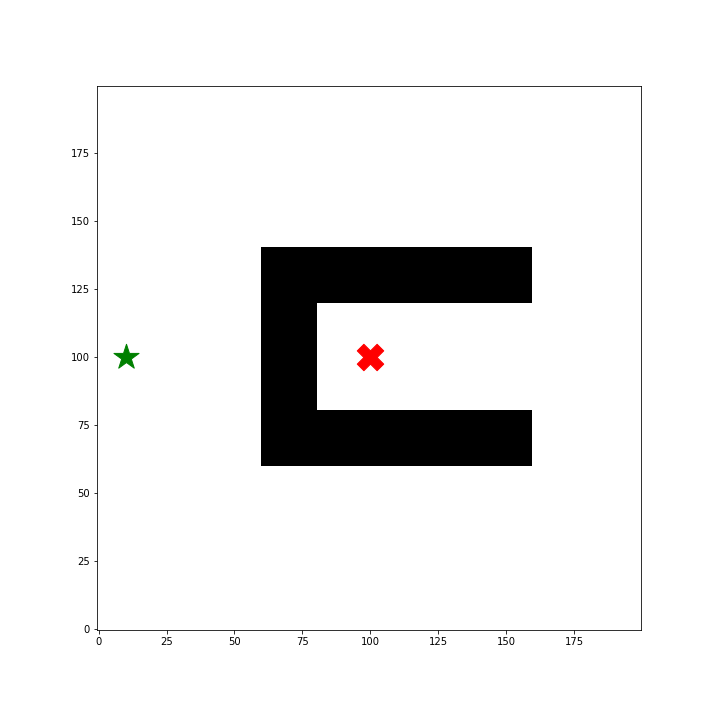}
	}
	\subfigure[Map 3]{
		\label{fig.map.map3}
		\includegraphics[width=0.3\linewidth]{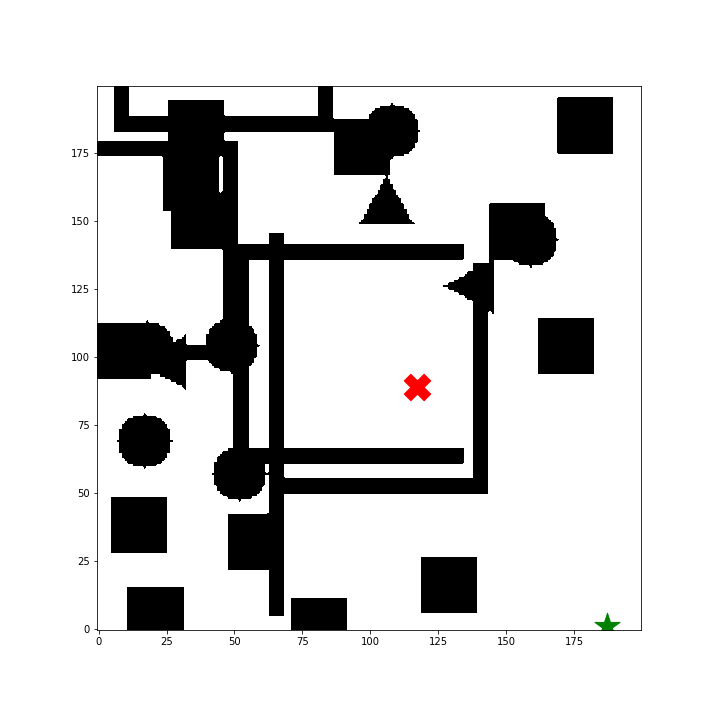}
	}
	\subfigure[Map 4]{
		\label{fig.map.map4}
		\includegraphics[width=0.3\linewidth]{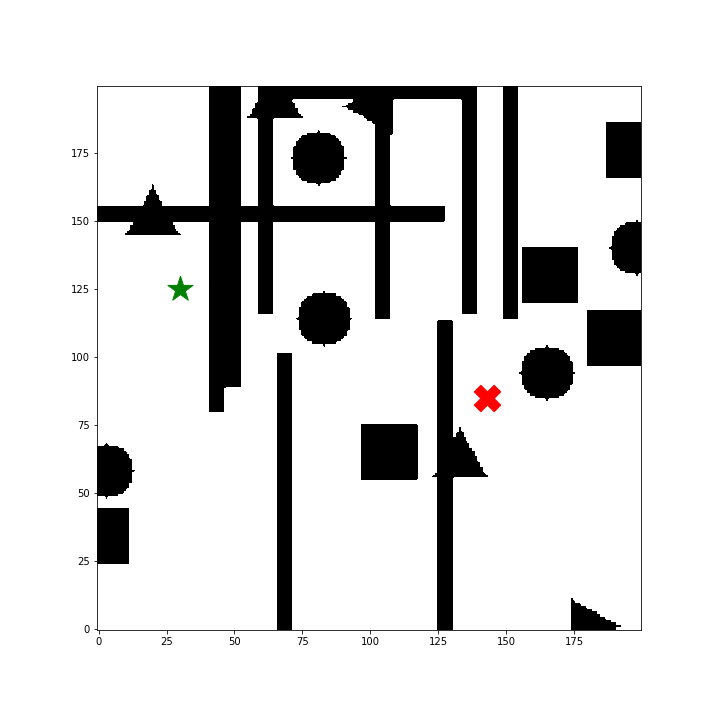}
	}
	\subfigure[Map 5]{
		\label{fig.map.map5}
		\includegraphics[width=0.3\linewidth]{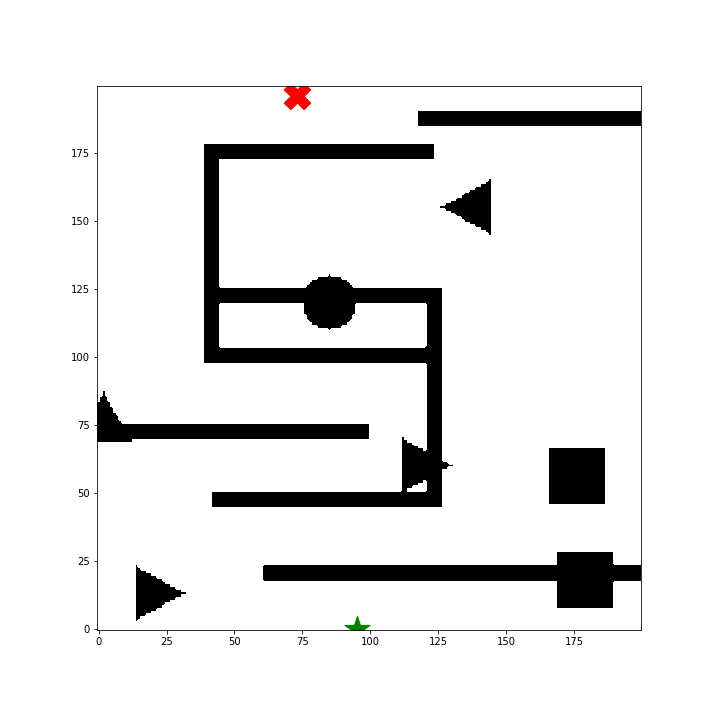}
	}
	\caption{Five floor maps with different complexities.  In each
          map, green star denotes the start state and the red cross is
          the destination.  }
	\label{fig.map}
	\end{figure}

{\bf Data} We generated a total of 10,000 maps for our
experiments. Each map consists of 12 randomly chosen start points and
12 randomly chosen goal points, resulting in a dataset of 1,440,000
maps in total.  Fig.~\ref{fig.map} shows five maps with different
levels of complexity. In our experiments, $70\%$ of the maps were used
for training, and $30\%$ for testing.

To ensure the simulation of diverse map scenarios, our map generator
randomly selects a variable number of obstacles, including shapes like
triangles, circles, squares, bars, and U-shaped obstacles. The
orientation of each obstacle is also chosen randomly.
Each map has a dimension of 200x200 pixels, and integer values are
assigned to each pixel based on their respective classes. Traversable
space is denoted by 0, obstacles are represented by 1, start points
are labeled as 2, and goal points are labeled as 3.

%In the next step, we utilize the A* algorithm to compute the optimal
%path for each map. Then, we extract the vertex points from the optimal
%path using our defined vertex point criterion. Our ground truth for
%model training is a binary mask indicating whether each pixel
%represents a vertex or not.

%The training process is conducted on a
%single NVIDIA 1080TI GPU using the PyTorch framework.

        \subsection{Sampling Probability Comparisons}
        
        The key design of our VertexNet RRT* lies in the modification
        of the training objective to focus on the turning points of
        the optimal paths. This modification is aimed to reduce the
        sampling space required for the RRT* algorithm.  By comparing
        the resulting probability distribution predictions from models
        trained with distribution predictions from models trained with
        the vertex-based objective versus the path-based objective, we
        can assess the effectiveness of our approach.

\begin{figure}[htb]
	\centering
	\subfigure[Optimal path]{
		\label{fig.prob.path}
		\includegraphics[width=0.45\linewidth]{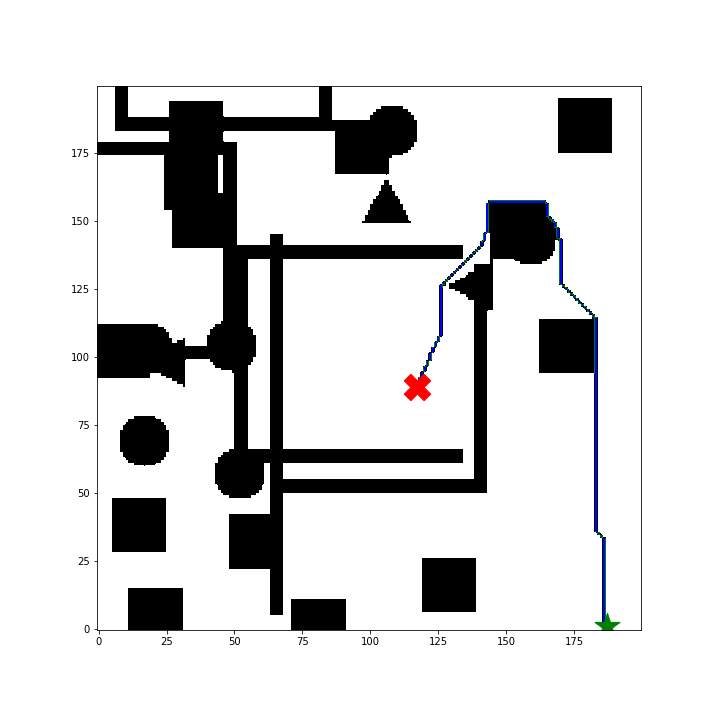}
	}
	\subfigure[Vertices in optimal path]{
		\label{fig.prob.vertex}
		\includegraphics[width=0.45\linewidth]{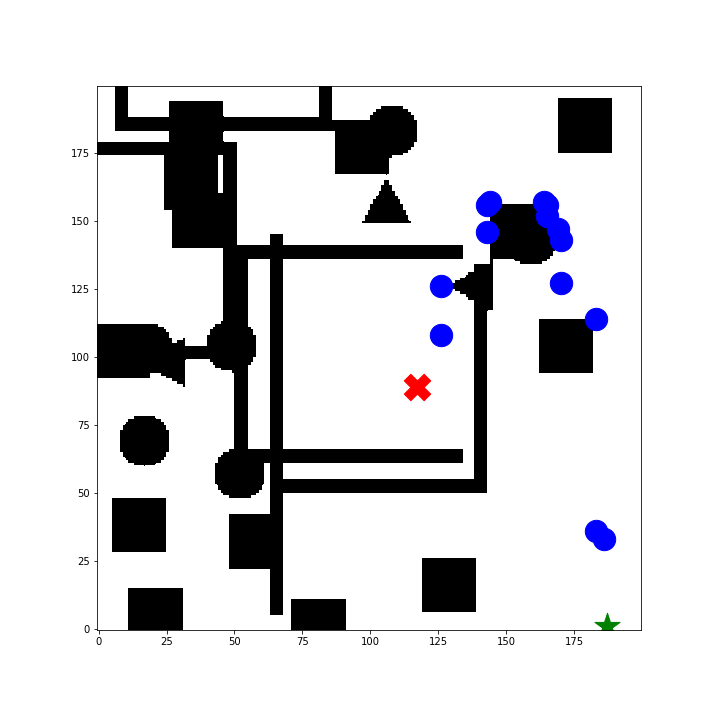}
	}	\subfigure[Neural RRT*]{
		\label{fig.prob.neural}
		\includegraphics[width=0.45\linewidth]{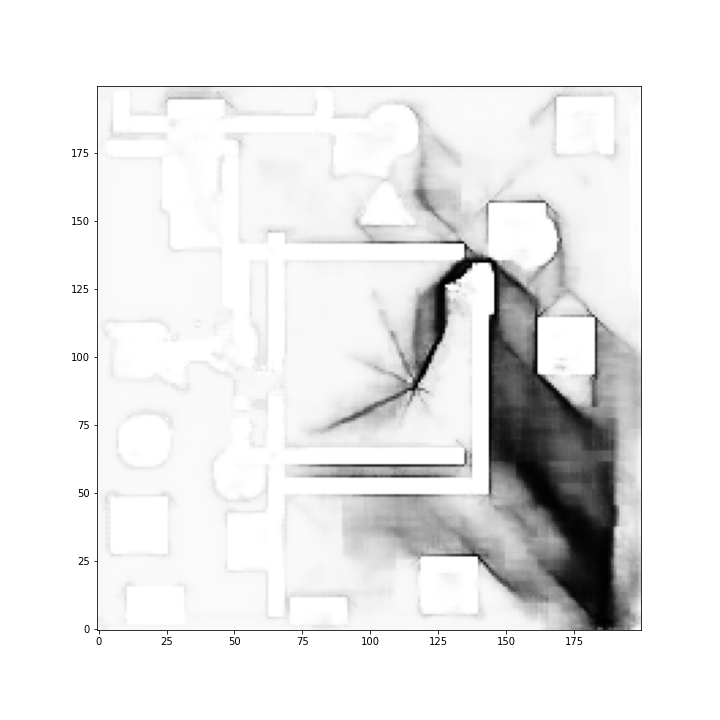}
	}
	\subfigure[VertexNet RRT*]{
		\label{fig.prob.vertexnet}
		\includegraphics[width=0.45\linewidth]{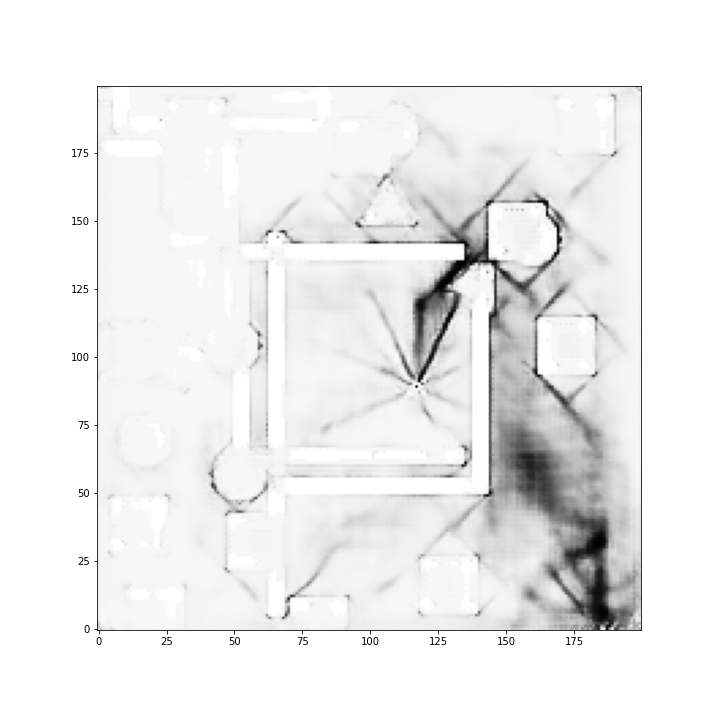}
	}
	\caption{Sampling distributions of Neural RRT* and VertexNet
          RRT*. Dark area depicts high probabilities, light area shows
          low probabilities.  (a) Blue line is the optimal path. (b)
          Blue dots are the vertices in the optimal path.  (c) Sampling
          probability distribution from Neural RRT*. (d) Sampling
          distribution from our VertexNet RRT*.  }
	\label{fig.prob}
	\end{figure}

        Fig.~\ref{fig.prob} illustrates a representative example,
        where the optimal path and the extracted vertices are
        displayed in Fig.~\ref{fig.prob}(a) and (b), respectively.
        Fig.~\ref{fig.prob}(c) shows the probability distributions
        trained using the optimal paths in Neural RRT*, while
        Fig.~\ref{fig.prob}(d) show the probability distributions
        trained using our VertexNet. As evident, the latter has much
        fewer bright areas, indicating that the sampling space of
        VertexNet RRT* is greatly reduced compared to the Neural RRT*.
        %achieving our design goal.
%   
%        trained using the whole optimal path (Fig.~
%        \ref{fig.prob.neural}) has more bright areas than the one
%        trained using only the verteices (Fig.~
%        \ref{fig.prob.vertexnet}).
        %This indicates that the sampling space of our VertexNet RRT*
        %is reduced compared to the Neural RRT*.
        % Overall, these results
        % demonstrate that we have achieved our design goal to
        %significantly reduce the sampling space required to construct
        %the optimal path.

\subsection{Path Planning Results and Analysis}

We performed experiments on four different algorithms, namely RRT*,
Neural RRT*, our VertexNet RRT* and Masked VertexNet RRT*. We use the
abbreviations RRT*, NRRT*, VNRRT* and M-VNRRT* in the upcoming
presentation. The evaluations were conducted for both {\it initial
  solutions}, where each algorithm terminates upon reaching the
destination, and {\it optimal solutions}, where each algorithm
terminates after finding the optimal path.

%We conducted initial experiments on the five individual maps in
%Fig~\ref{fig.map}, as well as 1,000 random maps. The 1,000 maps were
%randomly selected from the test set.  For each individual map, 1000
%trials are carried out for each algorithm on finding an initial
%solution. The experimental results for random maps are averaged across
%1000 maps. The Vertex Net RRT* is the fastest algorithm to find the
%initial solution in Random Maps settings as depicted in Table
%\ref{table.initial_solution_time_cost}. Moreover, the path length was
%comparable to that of the NRRT*.

We conducted experiments to find {\it initial solutions} on the five
individual maps in Fig.~\ref{fig.map}, as well as 1,000 random maps
selected from the test set. Each individual map underwent 1,000 trials
for each algorithm to find an initial solution. The results for the
random maps were averaged across the 1,000 maps. The performance of
the algorithms was evaluated based on {\it Path Length} and {\it Time
  Cost}, as summarized in
Table~\ref{table.initial_solution_path_length} and
Table~\ref{table.initial_solution_time_cost}. Among the models,
our VNRRT* algorithm demonstrated the fastest performance in
finding the initial solution in the Random Maps settings, as indicated
in Table~\ref{table.initial_solution_time_cost}. Moreover, the path
length achieved by VNRRT* was comparable to that of the baseline
Neural RRT*.

	\begin{table*}[!htb]
	
	\caption{Path Length comparisons of the {\it initial solutions}}
	\label{table.initial_solution_path_length}
	\centering
	\resizebox{.8\linewidth}{!}{
	\begin{threeparttable}
			\begin{tabular}{c|c|c|c|c|c|c}
				\hline
				& Map 1 & Map 2 & Map 3 & Map 4 & Map 5 & Random Maps \\
				\hline
				RRT* & 250.93$\pm$46.98 & 322.02$\pm$24.06 & 284.38$\pm$19.93 & 222.33$\pm$26.05 & 373.03$\pm$28.89 & 153.07$\pm$83.77 \\
				NRRT* & 214.49$\pm$47.46 & 316.9$\pm$26.02 & \textbf{282.62$\pm$18.39} & 210.23$\pm$18.8 & 364.38$\pm$30.6 & 142.71$\pm$80.65 \\
				VNRRT* & 215.85$\pm$47.96 & 312.17$\pm$23.2 & 283.94$\pm$18.57 & 212.53$\pm$19.55 & 361.57$\pm$30.98 & 145.75$\pm$81.38 \\
				M-VNRRT* $\tau$=0.5 & 164.18$\pm$35.44 & 304.33$\pm$23.00 & 293.17$\pm$21.17 & 205.7$\pm$17.52 & 351.48$\pm$32.01 & 134.25$\pm$79.92 \\
				M-VNRRT* $\tau$=0.9 & \textbf{154.02$\pm$24.01} & 313.67$\pm$25.77 & 296.76$\pm$22.96 & 202.30$\pm$16.42 & \textbf{344.03$\pm$29.34} & \textbf{133.98$\pm$80.73} \\
				M-VNRRT* $\tau$=0.99 & 163.62$\pm$32.51 & \textbf{302.72$\pm$23.08} & 306.46$\pm$29.52 & \textbf{198.09$\pm$18.85} & 346.61$\pm$24.52 & 136.55$\pm$82.49 \\
				Optimal & 135.14 & 255.68 & 264.71 & 190.20 & 319.58 & 119.35 \\
				
				\hline
				
			\end{tabular}	
		\begin{tablenotes}
			\footnotesize
			\item[] Optimal: Optimal path length
		\end{tablenotes}
	\end{threeparttable}
	}
	\end{table*}

	\begin{table*}[!htb]
	
	\caption{Time Cost of finding {\it initial solutions}}
	\label{table.initial_solution_time_cost}
	\centering
	\resizebox{.8\linewidth}{!}{
	\begin{threeparttable}	
			\begin{tabular}{c|c|c|c|c|c|c}
				\hline
				& Map 1 & Map 2 & Map 3 & Map 4 & Map 5 & Random Maps \\
				\hline
				RRT* & 0.19$\pm$0.20 & 0.7$\pm$0.52 & 33.42$\pm$46.82 & 6.76$\pm$13.47 & 8.06$\pm$11.09 & 1.84$\pm$14.72 \\
				NRRT* & 0.23$\pm$0.24 & \textbf{0.53$\pm$0.38} & 23.24$\pm$28.81 & 4.57$\pm$9.54 & 6.93$\pm$8.9 & 0.77$\pm$6.07 \\
				VNRRT* & 0.18$\pm$0.19 & 0.64$\pm$0.44 & 11.15$\pm$14.83 & 2.29$\pm$3.57 & 5.69$\pm$7.23 & \textbf{0.45$\pm$2.17} \\
				M-VNRRT* $\tau$=0.5 & 0.15$\pm$0.17 & 0.62$\pm$0.43 & 5.56$\pm$5.83 & \textbf{1.20$\pm$1.63} & 6.22$\pm$8.36 & 0.85$\pm$5.09 \\
				M-VNRRT* $\tau$=0.9 & \textbf{0.10$\pm$0.08} & 0.76$\pm$0.53 & \textbf{5.22$\pm$6.82} & 1.83$\pm$2.49 & 8.81$\pm$11.89 & 0.62$\pm$3.60 \\
				M-VNRRT* $\tau$=0.99 & 0.20$\pm$0.19 & 0.88$\pm$0.83 & 10.36$\pm$32.27 & 3.56$\pm$4.77 & \textbf{4.58$\pm$4.38} & 1.44$\pm$12.27 \\
				
				\hline
				
			\end{tabular}	
		\begin{tablenotes}
			\footnotesize
			\item[] Time cost to find the initial solution in seconds.
		\end{tablenotes}
	\end{threeparttable}
	}
	\end{table*}

%	\begin{figure*}[!htb]
%	\centering
%	\subfigure[Versus RRT*]{
%		\label{fig.initial_solution_time_improvement_vs_rrtstar}
%		\includegraphics[width=\linewidth, height=4cm]{Initial Solution Time Improvement VS RRT_}
%	}
%	\caption{Initial solution time improvement versus RRT* and NRRT* in percentage. (a) Time improvement of different algorithms compared with RRT*. (b) Time improvement of different algorithms compared with NRRT*.}
%	\label{fig.initial_solution_time_improvement}
%	\end{figure*}

	\begin{table*}[!htb]
	
	\caption{Time Cost of finding {\it optimal solutions}}
	\label{table.optimal_solution_time_cost}
	\centering
	\resizebox{.8\linewidth}{!}{
	\begin{threeparttable}	
			\begin{tabular}{c|c|c|c|c|c|c}
				\hline
					& Map 1 & Map 2 & Map 3 & Map 4 & Map 5 & Random Maps \\
				\hline
				RRT* & 1034.98$\pm$1055.17 & 50.97$\pm$35.56 & 47.20$\pm$57.36 & 28.55$\pm$17.03 & 36.19$\pm$23.21 & 29.62$\pm$102.66 \\
				NRRT* & 36.48$\pm$37.57 & 16.45$\pm$8.27 & 35.28$\pm$30.45 & 9.62$\pm$10.92 & 19.63$\pm$14.10 & 19.38$\pm$62.62 \\
				VNRRT* & 34.57$\pm$33.43 & 17.34$\pm$8.36 & 18.24$\pm$15.00 & 5.99$\pm$3.67 & \textbf{15.09$\pm$10.51} & 10.96$\pm$62.67 \\
				M-VNRRT* $\tau$=0.5 & \textbf{2.46$\pm$2.21} & \textbf{13.41$\pm$7.39} & \textbf{16.14$\pm$9.63} & \textbf{2.88$\pm$3.12} & 16.12$\pm$12.02 & \textbf{3.71$\pm$11.58} \\
				M-VNRRT* $\tau$=0.9 & 4.86$\pm$7.95 & 19.09$\pm$12.30 & 21.97$\pm$13.28 & 2.89$\pm$2.13 & 15.37$\pm$14.86 & 18.40$\pm$96.41 \\
				M-VNRRT* $\tau$=0.99 & 29.4$\pm$65.28 & 46.14$\pm$46.16 & 36.77$\pm$40.06 & 4.95$\pm$4.48 & 19.68$\pm$26.66 & 85.40$\pm$400.62 \\
				
				\hline
				
			\end{tabular}	
		\begin{tablenotes}
			\footnotesize
			\item[] Time cost to find the optimal solution in seconds.
		\end{tablenotes}
	\end{threeparttable}
	}
	\end{table*}

%\begin{figure}[!htb]
%	\centering
%	\includegraphics[width=\linewidth, height=4cm]{Optimal Solution Time Improvement VS RRT_ map1}
%	\caption{Optimal solution time improvement versus RRT* of Map 1 in percentage.}
%	\label{fig.optimal_solution_time_improvement_map1}
%\end{figure}
 
\begin{figure*}[!htb]
	\centering
	\resizebox{.8\linewidth}{!}{
    \includegraphics[width=\linewidth]{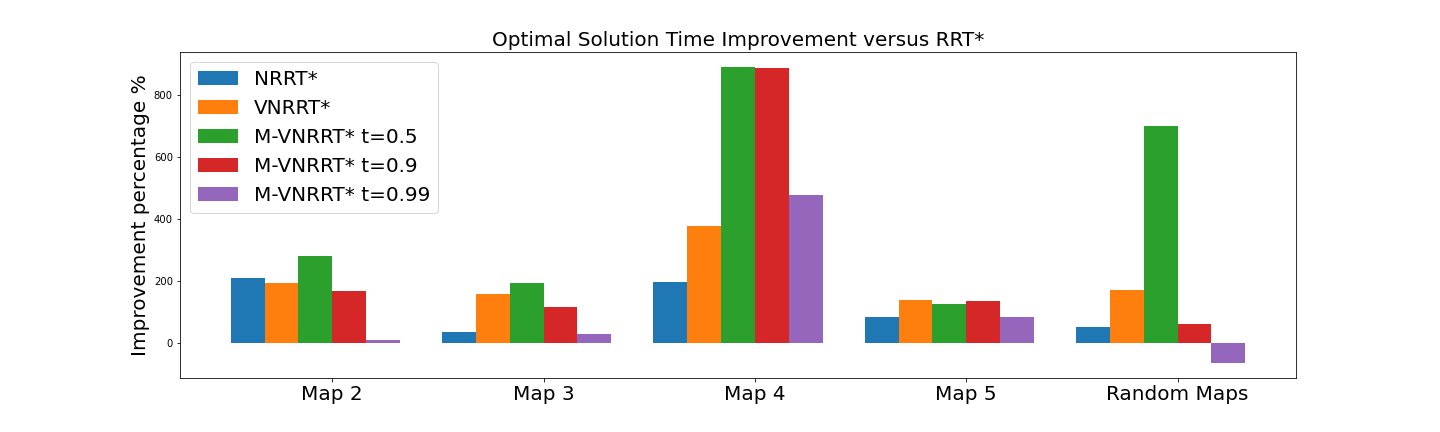}
    }
	\caption{Time improvements over RRT* on finding {\it optimal
            solutions}. Data are Maps 2-5 and Random Maps and the
          improvements are shown in percentage.}
	\label{fig.optimal_solution_time_improvement_restmaps}
\end{figure*}

%We also evaluate the performance of various algorithms in finding the
%optimal solution. Since finding the optimal solution is more
%time-consuming than the initial solution, we reduce the experimental
%trials to 100.
%
%Table \ref{table4} displays the number of iterations required to find
%the optimal path. Our Masked Vertex Ner RRT* $\tau$=0.5 algorithm
%outperformed the other algorithms in most experiments, with the
%exception of Map 5. The Vertex Net RRT* algorithm demonstrated better
%performance compared to the NRRT* algorithm across experiments
%on Map 3 to Map 5 and Random Maps. However, its number of iterations
%are still lower than those of the RRT* algorithm.
%
%We implemented a termination criterion in which the simulation
%concluded when the path length differed by no more than 1\% from the
%optimal path. As a result, the path length was almost identical
%across all experiments.
%
%The Masked Vertex Net RRT* $\tau$=0.5 algorithm delivered the best
%results across experiments on Map 1-4 and Random Maps, while the
%Vertex Net RRT* algorithm achieved the best performance on Map 5, as
%presented in Table \ref{table.optimal_solution_time_cost}.

We further evaluated the performance of the algorithms in finding {\it
  optimal solutions}. Due to the time-consuming nature of finding
optimal solutions, we reduced the number of experimental trials to
100. Among the algorithms tested, the M-VNRRT* $\tau$=0.5 algorithm
consistently delivered the best results in experiments conducted on
Map 1-4 and Random Maps. On the other hand, our VNRRT* algorithm
achieved the best performance on Map 5, as summarized in
Table~\ref{table.optimal_solution_time_cost}.

%To provide better insight into the relative convergence speed
%improvements, we have presented the visualization in Figure
%\ref{fig.optimal_solution_time_improvement_map1} and Figure
%\ref{fig.optimal_solution_time_improvement_restmaps}.

Fig~\ref{fig.optimal_solution_time_improvement_restmaps} shows the
speed improvements of the algorithms over RRT* on finding optimal
solutions.
%All networks-based algorithms demonstrated improvements,
%except for the Masked Vertex Net RRT* $\tau$=0.99 in the Random Maps
%experiment, possibly due to
%% its excessive masking of potential vertex points of the optimal path.
%its masking of very certain vertex points.
%
Our VNRRT* outperformed NRRT* in almost all the experiments, except
for Map 2, where its performance was 5\% worse. However, the M-VNRRT*
demonstrated relative improvements compared to NRRT* in all
experiments, particularly in Map 3, Map 4, and Random Maps, where the
improvements were 118.59\%, 234.03\%, and 422.37\%, respectively.

%Although algorithm performance may differ
%across different maps,
%As models'performance may differ for individual maps, the Random Maps
%experiment is the most objective, as it represents the average results
%of 100 maps. On average, the proposed VertexNet shows a
%76.82\% improvement over its baseline NRRT* algorithm when
%converging to the optimal path. The Masked Vertex Net RRT* $\tau$=0.5
%demonstrated a 422.37\% improvement when compared with NRRT*.

As models' performance vary on individual maps, the Random Maps
experiment provides a more objective evaluation by averaging results
of 100 maps. The results are summarized in the rightmost column of
Table 3. On average, the proposed VNRRT* algorithm shows an
acceleration of 76.82\% over the baseline NRRT* algorithm when
converging to optimal paths. The M-VNRRT* with $\tau$=0.5 demonstrates
an impressive improvement of 422.37\% when compared to NRRT*.

%Overall, VNRRT* has more than 70\% improvement over Neural
%RRT* when converging to both the optimal and initial solutions. The
%M-VNRRT* $\tau$=0.5 exhibits superior performance when
%converging to the optimal solution, being over 400\% faster than
%%NRRT*. However, it is 9.41\% slower than NRRT* when
%finding initial solutions.

In summary, our VNRRT* algorithm demonstrates a significant
improvement of over 70\% compared to NRRT* when converging to both the
optimal and initial solutions. On the other hand, our M-VNRRT*
$\tau$=0.5 exhibits superior performance in finding the optimal
solution, being over 400\% faster than NRRT*. However, it is 9.41\%
slower than NRRT* when finding initial solutions.

%VNRRT* is the fastest algorithm to find initial solutions (
%Table 2), perhaps due to the absence of the mask, we have less
%restriction for the sampling points to lie in the optimal vertices. In
%contrast, the mask puts more restriction for the sampling probability
%to draw from the optimal vertices, which speeds up the convergence to
%the optimal solution. This is supported by the Masked Vertex Net RRT*
%$\tau$=0.5 in optimal solution experiments. The points with
%probability prediction less than 50\% are unlikely to be vertex points
%but still have a probability of being sampled. Masking may have
%excluded these points, resulting in faster convergence to optimal
%solutions.

Our VNRRT* algorithm demonstrates superior speed in finding
initial solutions (Table 2), possibly because it does not incorporate
a mask that restricts the sampling points to optimal vertices. In
contrast, the inclusion of a mask in the M-VNRRT*
algorithm imposes more constraints on the sampling probability,
therefore speed up convergence to the optimal solution. This is
evident from the performance of M-VNRRT* $\tau$=0.5 in
the optimal solution experiments (Table 3). Points with a probability
prediction below 50\% are less likely to be vertex points but still
have a chance of being sampled. The masking process may have excluded
these points, resulting in faster convergence to the optimal
solutions.

%% file: con_liu.tex
\section{Conclusion}
In this work, we present a novel approach to improve the learned
heuristic for path planning algorithms. Rather than using the entire
optimal path line segment as the objective, we focus solely on the
turning points of the optimal paths in our proposed VertexNet. This
modification significantly enhances the speed of the path planning
algorithms.
%To address the
%imbalanced data problem caused by the vertex objective, we employ the
%Focal Loss technique in our deep learning model.
We also introduce a mask to the sampler of VertexNet RRT*, which
boosts the sampling probability of the actual vertices in the optimal
path. This further accelerates the convergence speed to the optimal
path.  Overall, our approach provides a more efficient method for
learning heuristics, which is important for future machine learning
applications in path planning problems.